\documentclass[10pt,twocolumn,letterpaper]{article}

\usepackage[pagenumbers]{cvpr} 

\usepackage{fontawesome}
\usepackage{booktabs}
\usepackage{tabularx}

\usepackage[dvipsnames]{xcolor}
\usepackage{algorithm} 
\usepackage{algpseudocode}

\definecolor{cvprblue}{rgb}{0.21,0.49,0.74}
\usepackage[pagebackref,breaklinks,colorlinks,citecolor=cvprblue]{hyperref}

\def\paperID{22} 
\def\confName{CVPR}
\def\confYear{2024}

\title{Red-Teaming Segment Anything Model}

\author{
  Krzysztof Jankowski$^{1,}$\thanks{equal contribution} \quad  Bartlomiej Sobieski$^{2,}$\footnotemark[1] \quad Mateusz Kwiatkowski$^{1}$ \\Jakub Szulc$^{1}$\quad Michał Janik$^{1}$\quad Hubert Baniecki$^{1}$\quad Przemysław Biecek$^{1, 2}$\\[0.5em]
  $^{1}$University of Warsaw \quad $^{2}$Warsaw University of Technology\\
  \tt\small \{kj.jankowski3, b.sobieski\}@student.uw.edu.pl \\ 
}

\begin{document}
\maketitle

\begin{abstract}
Foundation models have emerged as pivotal tools, tackling many complex tasks through pre-training on vast datasets and subsequent fine-tuning for specific applications. The Segment Anything Model is one of the first and most well-known foundation models for computer vision segmentation tasks. This work presents a multi-faceted red-teaming analysis that tests the Segment Anything Model against challenging tasks: (1) We analyze the impact of style transfer on segmentation masks, demonstrating that applying adverse weather conditions and raindrops to dashboard images of city roads significantly distorts generated masks. (2) We focus on assessing whether the model can be used for attacks on privacy, such as recognizing celebrities’ faces, and show that the model possesses some undesired knowledge in this task. (3) Finally, we check how robust the model is to adversarial attacks on segmentation masks under text prompts. We not only show the effectiveness of popular white-box attacks and resistance to black-box attacks but also introduce a novel approach - Focused Iterative Gradient Attack (FIGA) that combines white-box approaches to construct an efficient attack resulting in a smaller number of modified pixels. All of our testing methods and analyses indicate a need for enhanced safety measures in foundation models for image segmentation.
\end{abstract}

\section{Introduction}
\label{sec:intro}

The emergence of foundation models \cite{bommasani2022opportunities,radford2019language,brown2020language} rapidly changed the landscape of the applications of artificial intelligence~\cite{zhou2023foundation,ma2024segment}. Instead of training task-specific models from scratch, foundation models are trained on general data distributions using massive datasets and computing resources. This universal approach leads to emergent capabilities that allow for solving many challenging tasks. Foundation models can also be easily fine-tuned with domain-specific data, greatly improving the performance in complex tasks without losing previously acquired knowledge \cite{touvron2023llama,hu2022lora}. However, with great power comes great responsibility, and foundation models are no exception. A highly performant model must be deeply understood so that its limitations are known prior to using it in the real world~\cite{rombach2022high,shan2023glaze}.

Red-Teaming has recently become an integral part of foundation model deployments \cite{touvron2023llama}. This research field aims to evaluate the model from various points of view in order to detect any undesirable biases, inabilities, failure cases, and misalignments \cite{perez2022red}. It also proposes a variety of tools to fix the discovered drawbacks \cite{gao2023adaptive}. Performing a thorough Red-Teaming analysis is difficult, especially in the case of foundation models, as current machine learning algorithms achieve incomprehensible levels of complexity. Every attempt is doomed to shortcomings but provides important insights into possible means of improvement.

Motivated by these observations, we aim to perform a wide-ranging Red-Teaming analysis of the Segment Anything Model (SAM)~\cite{SAM} -- a foundation model for the segmentation task. Since its release, SAM has been already found useful in various tasks, e.g. medical imaging~\cite{medical_sam}. However, SAM's limitations are still not adequately studied, therefore our goal is to provide novel universal methods that can be applied to testing SAM and other segmentation models especially focusing on robustness, privacy, and vulnerability to adversarial attacks. Specifically,

\begin{figure*}[t]
  \centering
    \includegraphics[width=0.95\linewidth, trim=0cm 12.5cm 0cm 0cm, clip]{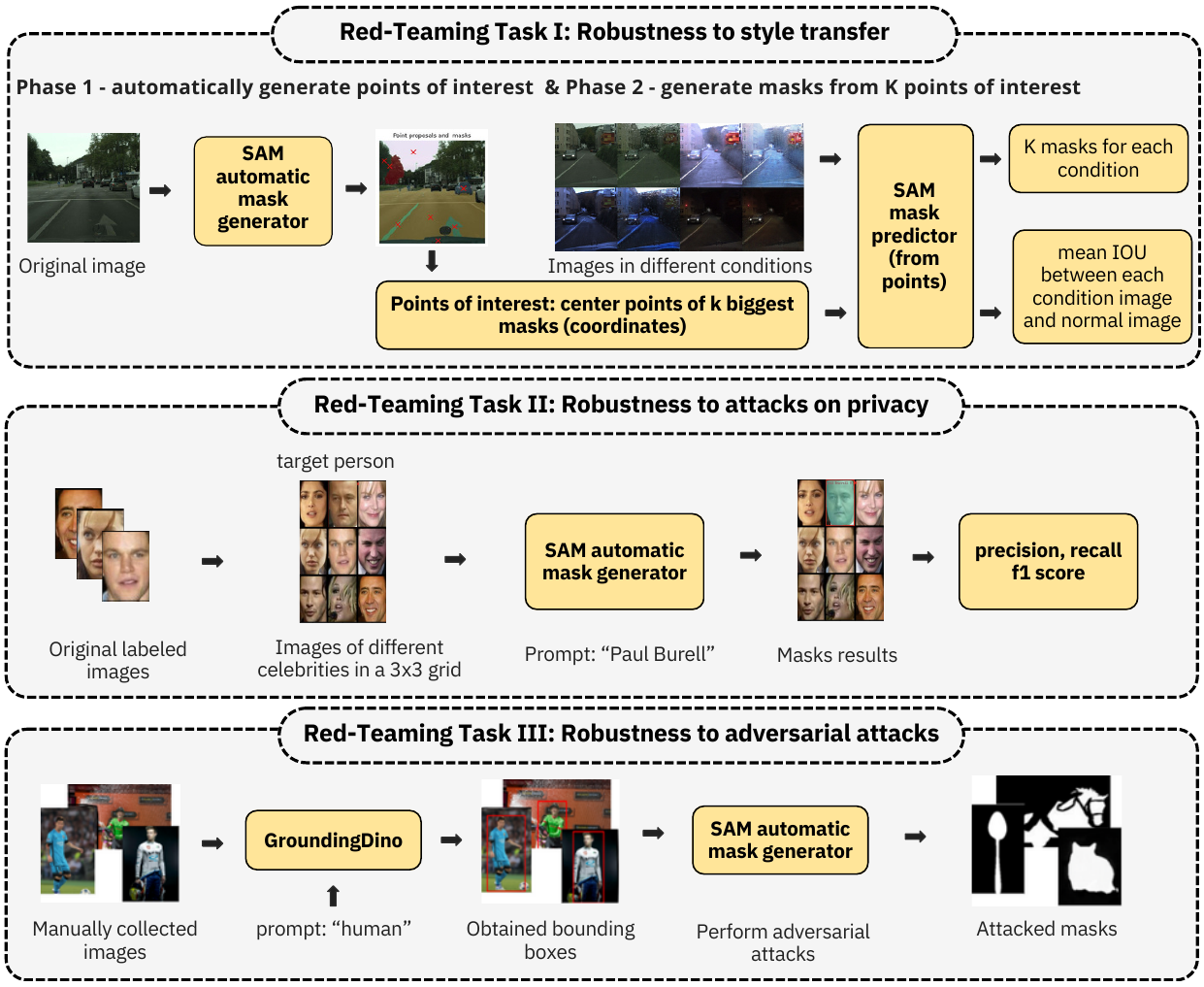}

   \caption{Overview of 3 Red-Teaming tasks on which Segment Anything Model is tested.}
   \label{fig:method1}
\end{figure*}

\begin{enumerate}
    \item We propose a novel style-transfer-based approach to verify SAM's robustness to different environmental conditions, showing that SAM is prone to realistic image perturbations. This analysis emphasizes that SAM should not be used as an out-of-the-box solution in critical practical scenarios like autonomous driving.
    \item We emphasize the importance of considering potential privacy security issues with SAM. Our findings reveal its capability to outperform a random classifier in facial classification tasks, highlighting concerns regarding its application in production-grade systems. This aspect necessitates careful consideration during the design phase of such systems and potential data leakage.
    \item We perform an extensive evaluation of SAM with a wide range of adversarial attacks. In addition to comparing white-box and black-box methods, we propose Focused Iterative
    Gradient Attack (FIGA) - a novel hybrid algorithm inspired by the observed vulnerabilities of SAM. Our approach makes the perturbations less noticeable while maintaining the attack's efficacy, further emphasizing the safety concerns.
\end{enumerate}

\noindent
We believe that our work may inspire a more wide-sighted view on further Red-Teaming of SAM, as well as other foundation models used in critical practical scenarios. We share our code on GitHub.\footnote{\url{https://github.com/JankowskiChristopher/red-teaming-segment-anything-model}}

\section{Related work}
\label{sec:related}

Segment Anything Model (SAM) \cite{SAM} is a foundation segmentation model consisting of an image encoder (pre-trained Vision
Transformer (ViT) \cite{vit}), flexible prompt encoder, and a transformer decoder. These building blocks enable prompting the model to generate masks from points of interest, bounding boxes, text prompts, or segment the whole image.

Thanks to great performance and ease of fine-tuning, SAM found use cases in many tasks ranging from image editing \cite{rombach2022high}, medical images segmentation~\cite{medical_sam}, or even intelligent vehicles \cite{sam_vehicles,sam_uav}. Due to being used in many critical areas, the model gained attention from researchers focusing on the robustness analysis. 

One popular type of analysis covers the model's robustness to different image augmentations~\cite{sam_noises} unveiling SAM's greater susceptibility to some distortions particularly Gaussian blur, radial blur, and chromatic aberration. This analysis gives more insight into the robustness of SAM, however, most of such distortions will never happen in real life, therefore we focus more on segmentation in style transfer setting \cite{corruptions_and_beyond}, especially motivated by another work \cite{texture_sam} showing that contrary to human vision, SAM is biased more towards texture rather than shape.

As prior work mostly focuses on changing the texture of the object which might lead to out-of-distribution data and happen extremely rarely in the real world, we shift our focus to a real-life scenario of different weather conditions in street photographs. Similar analyses were performed \cite{sam_weather_conditions} where adverse weather conditions were generated using Photoshop. However, these modified images are far from actually simulating real-world photos.

Evaluating the model explanation correctness is often performed with the use of bounding boxes that indicate areas on which the explanation should focus. However, in many practical scenarios, such boxes are not available. Merging images into grids is a simple technique that has proven useful in this context \cite{arias2022focus,rao2022towards}. It allows for creating synthetic images with user-defined regions that contain the complete knowledge required to perform a correct prediction. For example, a grid can be constructed from three images of dogs and one image of a cat and used as model input. If the animal classifier predicts the cat class, the model explanation should indicate only the cat image from the grid. Our work relates to this line of research by exploiting grids of images in red-teaming to test SAM's robustness to privacy attacks with no connection to other explanation techniques.

SAM has been also subject to different attacks including white-box \cite{attacksam}, black-box \cite{sam_blackbox}, and even Universal Adversarial Perturbation (UAP) \cite{uap} where a small perturbation resulted in degradation of mask quality in many images~\cite{sam_uap}.
Neural network-based models are particularly vulnerable to adversarial attacks \cite{nn_properties}, so in our work, we decided to test several approaches to find potential weaknesses in SAM. One of the primary algorithms in this field is the Fast Gradient Sign Method (FGSM) \cite{goodfellow2015explaining}, which uses gradients computed over input and, based on these, generates noise to disrupt the model performance. This method is not always immediately successful, so it can be used iteratively, introducing small perturbations \cite{pgd}. As this is a very popular algorithm, its effectiveness has already been tested on SAM \cite{attacksam}. We decided to make changes to the update rule to try to achieve any desired mask at a lower cost. However, always changing all pixels can sometimes be undesirable, so we also examined attacks on a limited area of the image. An extreme case of this approach is the Jacobian-based Saliency Map Attack (JSMA) \cite{jsma}, in which only one pixel is selected for attack.

The most versatile type of attacks are black-box attacks, as we can also apply them when we do not have access to the weights of a particular version of the model. All we need is the possibility of multiple inferences. Many tools are used in this field to find the right direction of attack, these include random orthonormal vectors \cite{simba} and surrogate models \cite{ebad}.

\section{Methods}

\subsection{Robustness to style transfer}
Robustness of SAM has been extensively tested on different perturbed images \cite{sam_robustness,sam_noises,sam_weather_conditions} however because these perturbations usually consist of artificial image filters, they are out of distribution data and won't appear in real-life scenarios. Therefore our analysis focuses on testing the robustness of segmentation masks in real-life scenario image perturbations.

To accomplish this goal, we test the quality of generated masks under different styles from the Multi-weather-city dataset \cite{MultiWeather} which is a modified version of the
Cityscapes dataset \cite{Cityscapes} consisting of dashboard images of streets in different cities. The original Cityscapes images were modified using a set of GAN \cite{gan} and CycleGAN \cite{cycle-gan} methods to obtain image variants under 7 generated adverse
weather conditions: night, snow, wet, and additional variants with drops on the windshield.

To reliably assess the quality of generated masks under different styles, we have to make sure that we correctly compare the masks among images. To achieve this goal,
as illustrated in Figure \ref{fig:method1} Task I, our
pipeline starts with original photos and uses SAM automatic mask generation mode
to generate all the masks for each image. We use a heuristic of picking $k$ largest
masks as they hold information about the most important objects and calculate the center point for
each mask. These points are treated as mask IDs to later reliably compare the corresponding masks.

In the second part, we take all the images with the changed style
and corresponding coordinates of $k$ points from the original image and we calculate the
masks using the SAM mask predictor from points mode. We obtain masks for each image and calculate the mean Intersection over Union (IoU) \cite{meanIOU} between the obtained masks and the original masks. We recognize that style transfer changes might lead to alterations in images, including changes in colors and potentially different masks. However, we consider these variations to be minor and not of significant concern.

\subsection{Robustness to attacks on privacy}

With the growing popularity of computer vision models in downstream tasks, potential privacy concerns have been raised \cite{privacy_concerns}. As SAM can be prompted with text prompts, we examine its capacity to
classify celebrity faces. This inquiry is crucial, as it underscores potential misuse implications associated with unintended classification.

Although SAM is open-source, textual prompting is not available in the original source code. Therefore, we
use LangSAM~\cite{langsam}, an open–source implementation that connects SAM with GroundingDINO~\cite{dino} which serves as a text encoder instead of originally used CLIP \cite{clip} making textual prompting possible. It also strongly shows how flexible the prompt encoder is and how easily it can be changed.

Our analysis is based on the CelebA dataset \cite{CelebA} with selected 16 different celebrities, evenly split by gender and from diverse backgrounds, to ensure broad representation in our study. As illustrated in Figure \ref{fig:method1} Task II, we initiate each iteration of the experiment by randomly selecting nine images of different celebrities' faces and putting them in a $3\times3$ grid. This deliberate choice of using multiple faces, as opposed to just one, prevents the model from ``cheating" by consistently segmenting the sole person in an image; thus, making this method more reliable. We then prompt the LangSAM model with the full name of the celebrity and as a result, obtain the indices of the grid in which the model thinks that the celebrity is located. We only consider indices with bounding boxes where a mask is present as valid.

We evaluate the task as a binary classification i.e., for each face in a grid, we determine whether it is a true positive, true negative, false positive, or false negative. Subsequently, we measure precision, recall, and the F1 score over multiple iterations. To see whether the model robustly classifies celebrities, we repeat the tests and permute faces within each grid 5 times.

We approach LangSAM as an end-to-end system, evaluating its ability to accurately classify celebrities while considering potential biases holistically.
 These biases could largely originate from the GroundingDINO prompt encoder. However, as we have previously argued, the flexible nature of the prompt encoder, which can be easily interchanged, underscores the importance of analyzing its performance to minimize risks when such a model is deployed.

\subsection{Robustness to adversarial attacks}

Adversarial attacks aim at producing an imperceptible perturbation in the image that forces the model to return clearly inaccurate outputs. Specifically, for a given model $f$ and image $\mathbf{x}$, adversarial attacks can be generally understood as solutions to the following optimization problem:
\begin{equation}
    \min_{\delta} - \Vert f(\mathbf{x}) - f(\mathbf{x} + \delta) \Vert + \lambda \cdot \Vert \delta \Vert, 
    \label{eq:opt_adv}
\end{equation}
i.e. we want to find the smallest perturbation $\delta$ that results in the biggest difference between the prediction before and after adding this perturbation to the image. In Equation \ref{eq:opt_adv}, $\lambda$ accounts for the trade-off between the norm of the difference of predictions and of the perturbation. In terms of \emph{targeted} attacks, we additionally assume that the perturbation should lead to a specified prediction of the model.

The methods for creating these perturbations can be generally categorized as \textit{white-box}, if they assume access to the model's weights and \textit{black-box} otherwise.

In this paper we focus on white-box methods: Fast Gradient Sign Method (FGSM) \cite{goodfellow2015explaining}, Jacobian-based Saliency Map Attack (JSMA) \cite{jsma} and black-box methods: Simple Black-box Adversarial Attacks (SIMBA) \cite{simba} and Ensemble-based Blackbox Attacks (EBAD) \cite{ebad}.

\subsubsection{White-box attacks}

FGSM operates by making uniform adjustments to all pixels in an image based on the gradient of the loss with respect to the input image. This method is known for its efficiency, in rapidly generating adversarial examples, but it can lead to overt modifications that are easier to detect due to the broad scope of its perturbations. In our approach, we modify the original loss to:
\begin{equation}
    X := X - \epsilon \cdot \text{sgn}(\nabla_X \|\sigma(\text{SAM}(p, X))-\text{Y}\|_2^2)
    \label{eq:fgsm2}
\end{equation}
where $X$ is the image, $Y$ is the target, $p$ is the text prompt, and the epsilon value for pixels in the range $[0,255]$ is set to $1$.
The modified formula is better suited for segmentation problems, as it can be useful in solving more sophisticated tasks.
We denote the modified algorithm as FGSM*.

On the other end of the spectrum, JSMA takes a more meticulous route than FGSM. It focuses on altering individual pixels one at a time, guided by a saliency map that ranks pixels based on their impact on the model's output. While this approach allows for more discreet perturbations, it is significantly slower and computationally more intensive than FGSM.

In order to reduce the shortcoming of the mentioned methods, we introduce a novel approach Focused Iterative Gradient Attack (FIGA). FIGA is designed to bridge the gap between these extremes. It targets the top $k$ pixels guided by the saliency map and adjusts their values by a fixed amount $\epsilon$ in the direction indicated by the gradient.

\begin{figure*}[h]
  \centering
  \includegraphics[width=0.9999\textwidth, trim=0 24.5cm 0 0cm]{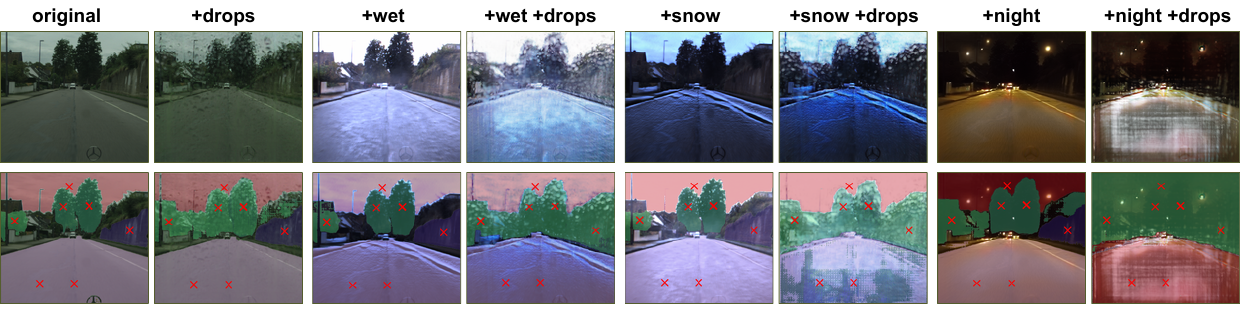}
  \caption{Example SAM predictions for the original image and with added different weather conditions. Respective masks have the same colors, but due to opacity and changing colors of the images the masks' colors change as well. Red crosses correspond to the points of interest for which the corresponding masks were generated.}
  \label{fig:masks}
\end{figure*}

FIGA allows for a tailored balance: at $k=1$ the method is equivalent to JSMA and at $k$ equal to the total number of pixels it mirrors FGSM. Choosing a $k$ between the two extremes achieves a strategic balance between efficiency and stealthiness. By targeting a select group of pixels rather than adjusting them all uniformly or individually, our method can generate adversarial examples that are less noticeable than those produced by FGSM but more quickly and with less computational cost than those generated by JSMA. For a pseudocode of a single iteration of our method, see \cref{code:attack}.

\begin{algorithm}
	\caption{FIGA iteration} 
	\begin{algorithmic}[1]
        \State \textbf{Input:}
        \State \emph{$X$, $Y$} \Comment{Input image and target mask}
        \State \emph{$v=\nabla_X J(f(X),Y))$} \Comment{gradient $v$ w.r.t. loss $J$}
        \State \emph{$k$} \Comment{Number of pixels to modify}
        
        \\
        \Function{FIGA}{$X$, $v$, $k$}
        \State $\text{topk} \leftarrow \operatorname{sorted}(|v|,\text{desc}))[k]$
        \For{$i=1,2,\ldots$}
            \If{$|v[i]| < \text{topk}$}
                \State $v[i] \leftarrow 0$
            \EndIf
        \EndFor
        \State $X \leftarrow X - \epsilon\cdot \operatorname{sign}(v)$\\
        \hspace*{\algorithmicindent} \Return $X$
        \EndFunction
	\end{algorithmic} 
    \label{code:attack}
\end{algorithm}

\subsubsection{Black-box attacks}

SIMBA operates as a black-box technique, employing iterative pixel perturbations with random orthonormal vectors. One of the components of SIMBA is the Discrete Cosine Transform (DCT) \cite{ahmed1974dct} that filters out high noise frequencies. This component proved effective in enhancing the method's accuracy against sophisticated models like SAM.

Transfer-based attacks use a group of similar surrogate models to create adversarial attacks on a black-box target model. The EBAD algorithm adapts this for image segmentation, enhancing the transferability of attacks from the surrogate models to the target. In EBAD, each iteration starts with generating a perturbation using the PGD \cite{pgd} method to deceive multiple surrogate models simultaneously, by optimizing a combined prediction score.

\section{Results}
\label{sec:results}

\subsection{Robustness to style transfer}

Thanks to its general architecture, our testing method can be applied to various images and style transfer techniques. However, we specifically utilize the Multi-weather-city~\cite{MultiWeather} dataset to evaluate the performance of systems using SAM for intelligent vehicles~\cite{sam_vehicles} across diverse road conditions which is crucial for deploying such systems safely.

\begin{figure*}[h]
\includegraphics[width=0.999\textwidth, trim={0cm 0.75cm 0cm 0cm}]{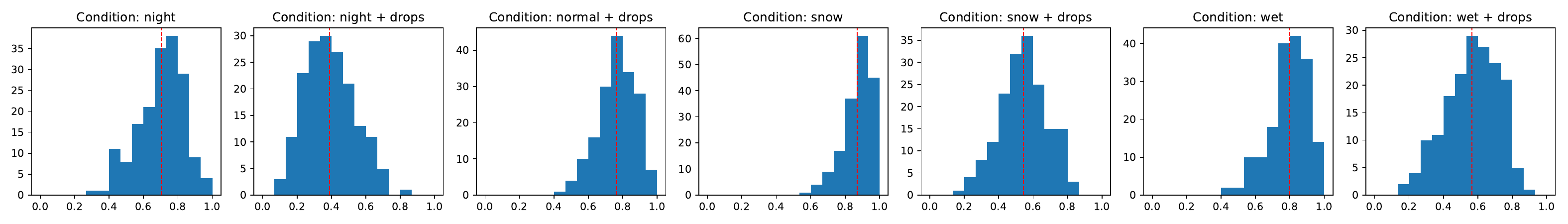}
   \caption{Distributions of mean IOUs between original image masks and augmented image masks for all weather conditions. Red vertical line represents the mean of the distribution. Exact values of mean and standard deviation are presented above the histograms.}
   \label{fig:histograms}
\end{figure*}

Figure \ref{fig:histograms} presents the distributions
of mean IOUs between masks for different weather conditions. Each histogram represents the results of the comparison between original images and images with the selected adverse condition applied. We can observe that when conditions are
not that harsh i.e. snow and night, the means of the distributions are 0.87 and 0.80 respectively. In wet conditions, the mean decreases to 0.70. When drops are present on the
windshield, the performance drops even further to 0.39 (night + drops) and masks often are
destroyed as can be seen in Figure \ref{fig:masks}. All distributions approximately follow the normal distribution and
the results align with intuition, showing that as the conditions deteriorate, the performance also declines.

We can further analyze different weather conditions by noticing that conditions: ``snow", ``night" and ``wet" predominantly change the colors but added conditions ``drops" influence the texture of the images. This aligns with research showcasing that contrary to intuition SAM is more biased towards texture rather than shape~\cite{texture_sam}. Therefore masks generated for images with ``drops" are significantly worse. 

As the presence of the drops on the images is so crucial, we believe that it has to be taken into account when designing algorithms for autonomous vehicles which might not work under heavy rain conditions posing serious risks.

\begin{table*}[h]
    \centering
    \label{tab:weather_conditions}
    \small
    \centering
    
    \begin{tabularx}{\textwidth}{*{7}{X}}
        \toprule
        night & night + drops & normal + drops & snow & snow + drops & wet & wet + drops \\
        \midrule
        0.70 $\pm$ 0.13 & 0.39 $\pm$ 0.14 & 0.76 $\pm$ 0.11 & 0.87 $\pm$ 0.08 & 0.54 $\pm$ 0.14 & 0.80 $\pm$ 0.11 & 0.57 $\pm$ 0.16 \\
        \bottomrule
    \end{tabularx}
\end{table*}

\subsection{Robustness to attacks on privacy}

The results in Table \ref{tab:text_prompts} demonstrate that the model’s performance varies significantly depending on the celebrity name used in the prompt.

\begin{table}[h]
    \centering
    \small
    \begin{tabular}{l@{\hspace{2pt}}r@{\hspace{2pt}}@{\hspace{2pt}}r@{\hspace{2pt}}@{\hspace{2pt}}r@{\hspace{2pt}}}
        \toprule
        \textbf{Celebrity \faFemale} & \textbf{Precision (\%)} & \textbf{Recall (\%)} & \textbf{F1 (\%)} \\ 
        \midrule
        Angelina Jolie & 22.9 $\pm$ 0.8 & 73.4 $\pm$ 2.4 & 34.9 $\pm$ 1.1 \\
        Britney Spears & 23.2 $\pm$ 0.5 & \textbf{96.6 $\pm$ 1.7} & 37.4 $\pm$ 0.8 \\
        Jennifer Aniston & 15.6 $\pm$ 1.5 & 40.7 $\pm$ 4.0 & 22.6 $\pm$ 2.2 \\
        Madonna & 11.2 $\pm$ 0.1 & \textbf{99.7 $\pm$ 0.5} & 20.1 $\pm$ 0.1 \\
        Salma Hayek & 17.3 $\pm$ 0.4 & 87.4 $\pm$ 2.4 & 28.8 $\pm$ 0.7 \\
        Jennifer Lopez & 24.9 $\pm$ 1.7 & 53.6 $\pm$ 3.9 & 34.0 $\pm$ 2.3 \\
        Keira Knightley & 25.0 $\pm$ 2.6 & 43.0 $\pm$ 4.9 & 31.6 $\pm$ 3.4 \\
        Nicole Kidman & 12.7 $\pm$ 0.8 & 66.0 $\pm$ 4.3 & 21.3 $\pm$ 1.3 \\
        \bottomrule
        \toprule
        \textbf{Celebrity \faMale} & \textbf{Precision (\%)} & \textbf{Recall (\%)} & \textbf{F1 (\%)} \\ 
        \midrule
        Keanu Reeves & 4.0 $\pm$ 1.2 & 8.7 $\pm$ 2.7 & 5.5 $\pm$ 1.7 \\
        Orlando Bloom & 11.1 $\pm$ 0.1 & \textbf{100.0 $\pm$ 0.0} & 20.0 $\pm$ 0.1 \\
        Prince William & 32.5 $\pm$ 2.0 & 68.0 $\pm$ 4.8 & \textbf{43.9 $\pm$ 2.8} \\
        Abdullah Gul & 25.7 $\pm$ 2.2 & 50.0 $\pm$ 4.4 & 33.9 $\pm$ 2.8 \\
        Paul Burrell & 34.3 $\pm$ 1.4 & 80.6 $\pm$ 3.6 & \textbf{48.1 $\pm$ 1.9} \\
        Mick Jagger & 11.9 $\pm$ 0.6 & 73.1 $\pm$ 3.9 & 20.4 $\pm$ 1.0 \\
        Michael Jackson & 11.1 $\pm$ 0.1 & \textbf{100.0 $\pm$ 0.0} & 20.0 $\pm$ 0.1 \\
        Nicolas Cage & 31.5 $\pm$ 2.7 & 52.1 $\pm$ 4.8 & 39.2 $\pm$ 3.4 \\
        \bottomrule
    \end{tabular}

    \caption{Precision, recall, and F1 scores for celebrity classification, accompanied by their respective standard deviations, in relation to grid cell permutations. Celebrities are divided into equally sized gender groups.}
    \label{tab:text_prompts}
\end{table}

\begin{figure}[h!]
  \centering
  \includegraphics[width=0.999\columnwidth, trim=1.5cm 21.8cm 1.5cm 0cm, clip]{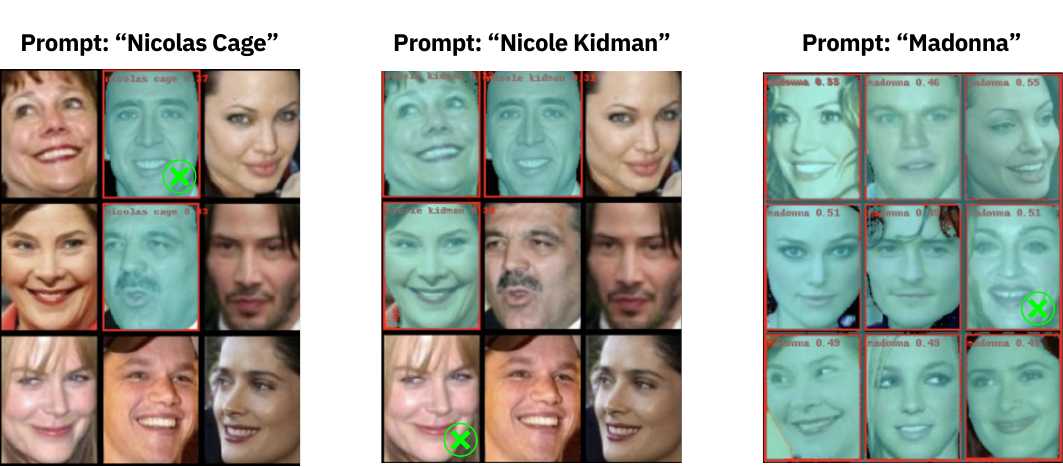}
  \caption{Example results of segmenting celebrity faces. Images are in a $3$x$3$ grid with a small green cross in the bottom-right corner of the ground-truth image for the text prompt above the grid. Colored images with a red bounding box are the model's answer to the text prompt. In the first and last example, the model correctly segmented the person but introduced false positives. In the middle example, the model did not correctly classify the person.}
  \label{fig:celeba_masks}
\end{figure}

Notably, certain celebrities, such as \emph{Prince William} and \emph{Paul Burrell}, achieve substantially higher precision, recall, and F1 scores, which is particularly intriguing given their association with the British royal family. Conversely, some celebrities, like \emph{Keanu Reeves}, exhibit scores markedly lower than those expected from a random classifier.

Additionally, the experiment reveals some unique anomalies: names such as \emph{Orlando Bloom}, \emph{Madonna}, and \emph{Michael Jackson} lead the model to provide less conservative predictions, frequently predicting all images in a grid as containing the target celebrity.

These results indicate that the model exhibits highly variable performance for different celebrity names, deviating significantly from the behavior expected of a random classifier. This underscores the nuanced understanding and bias inherent in the model, which deviates significantly from random chance in its predictions. Example masks for different celebrity names as prompts are depicted in Figure \ref{fig:celeba_masks}.

\subsection{Robustness to adversarial attacks}

To evaluate SAM's robustness to adversarial attacks in a practical and realistic scenario, we base our evaluation on a case study of segmenting single individuals when prompting SAM with only textual input. This setting aims at reflecting a simple use case of SAM when the user requires the model to provide accurate segmentation masks when provided with only a very general conditioning signal.

To simulate a real-life setting, we divide our approach into three distinct phases (see \cref{fig:method1} Task III):

\begin{enumerate}
    \item We first manually choose 100 images of single individuals from the original training dataset of SAM, which should give the model a slight advantage. 
    \item Using LangSAM, we simulate SAM's textual prompting. Each image is fed into GroundingDINO alongside a fixed descriptor (e.g., ``human'') to generate a bounding box, ideally encompassing the individual's area. Then, we input the image to the original SAM with the bounding box as a prompt. This design is due to the SAM limitation of not allowing textual prompting. To ensure that only SAM is attacked, we keep the initial bounding box fixed throughout the attack phase.
    \item We perform an adversarial attack with the goal of inverting the ground truth mask in the final segmentation. 
\end{enumerate}

The effectiveness of each algorithm was assessed with different textual prompts: \textit{person}, \textit{human}, and \textit{man}. For a qualitative comparison, we include the mean and standard deviation of IoU between the original and attacked mask, Mean-squared Error (MSE), and the infinity norm ($L_{\inf}$) of the difference between the original and the attacked image in Table \ref{tab:main_results} for each method and prompt.

\begin{table}[ht]
    \centering
    \small
    \begin{tabular}{@{\hspace{2pt}}lrrr@{\hspace{2pt}}}
    \toprule
         \textbf{Method} & \textbf{IoU (\%) $\downarrow$}  & \textbf{MSE $\downarrow$} & $\mathbf{L_{\inf}} \downarrow$ \\
         \midrule
         \multicolumn{4}{c}{prompt: \textit{person}} \\
         \midrule
         FGSM* & 1.4 $\pm$ 5.2 & 13 $\pm$ 11  & 11.6 $\pm$ 8.9 \\
         FIGA & 1.7 $\pm$ 2.2 & 7 $\pm$ 4 & 254.2 $\pm$ 2.35 \\
         SIMBA & 67.0 $\pm$ 23.0 & 2927 $\pm$ 1699  & 226.4  $\pm$ 45.7 \\
         EBAD & 96.0 $\pm$ 4.0 & 29549 $\pm$ 2403 & 247.6 $\pm$ 3.7 \\
         \midrule
         \multicolumn{4}{c}{prompt: \textit{human}} \\
         \midrule
         FGSM* & 5.7  $\pm$ 19.6  &  13  $\pm$ 11  &   11.6 $\pm$ 8.9 \\
         FIGA & 19.0 $\pm$ 23.4  &  7 $\pm$ 4  &  254.2 $\pm$ 2.4  \\
         SIMBA & 67.5  $\pm$ 23.8 & 2928  $\pm$ 1699   & 226.0  $\pm$ 45.7 \\
         EBAD & 94.0  $\pm$ 4.0 & 27322  $\pm$ 1849 & 243.2  $\pm$ 3.2 \\
         \midrule
         \multicolumn{4}{c}{prompt: \textit{man}} \\
         \midrule
         FGSM* & 3.3  $\pm$ 12.3  &  13 $\pm$ 11 &  11.6 $\pm$ 8.9 \\
         FIGA & 19.0 $\pm$ 24.3  & 7  $\pm$ 3  &  254.2  $\pm$ 2.4  \\
         SIMBA & 68.1 $\pm$ 22.9 & 2928  $\pm$ 1698  & 227.1  $\pm$ 46.1 \\
         EBAD & 93.0  $\pm$ 3.0 & 28199  $\pm$ 2132 & 244.3  $\pm$ 2.8 \\
         \bottomrule
    \end{tabular}
    \caption{Results for white-box and black-box attacks on SAM. The mean and standard deviation for the prompts: \textit{person}, \textit{human}, \textit{man} are presented.}
    \label{tab:main_results}
\end{table}

\subsubsection{White-box attacks}

We began our tests by comparing the white-box methods. Results indicate that our algorithm FIGA requires less perturbative noise to achieve comparable or superior adversarial effects when measured against modified FGSM, based on MSE comparisons. Additionally, our method outpaces JSMA in execution times, offering a more efficient alternative for generating adversarial examples.

Through extensive testing utilizing the Optuna framework for hyperparameter optimization \cite{optuna}, we identified optimal settings for FIGA at $k=2653$ and $\epsilon=5$. These parameters balance the effectiveness of adversarial perturbations with minimal noise introduction and expedited computation.

\begin{figure}[h]
\centering
\includegraphics[width=0.999\columnwidth, trim={2.5cm 24cm 2.5cm 0cm}, clip]{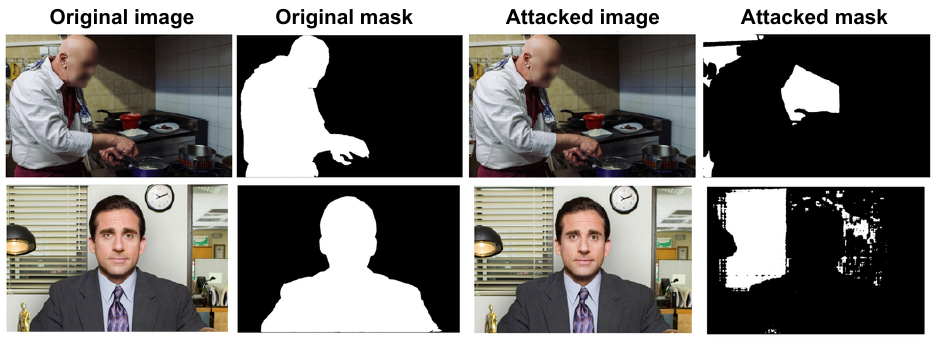}
   \caption{Examples of attacks with perturbations small enough to be imperceptible by the human eye, created using an FGSM-based approach. The attack successfully destroys the original masks.}
   \label{fig:fgsm_attacks}
\end{figure}

Our proposed method only makes changes on a limited number of pixels, so some of the changes can be very large, as can be seen in the norm $L_{\inf}$. In terms of this metric, our modified FGSM algorithm performs better, as it generates noise that is spread over the surface of the entire image. The differences between the original images and the generated attacks are invisible to the human eye, as shown in Figure~\ref{fig:fgsm_attacks}.

Initially, we generated an attack for each image in our dataset and prompt ``person'' and then we tested the resulting adversarial examples for other prompts (``human'' and ``man''). In this comparison the FGSM performed very well, demonstrating the robustness of this algorithm.

\begin{figure}[h]
\centering
\includegraphics[width=0.999\columnwidth, trim={2.5cm 24.5cm 2.5cm 0cm}, clip]{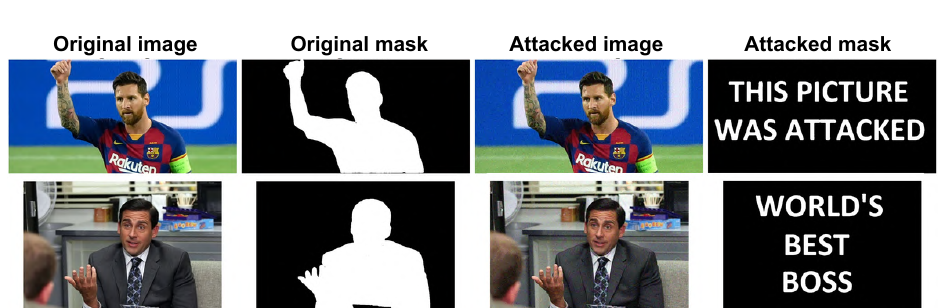}
   \caption{Examples of targeted attacks, generated using an FGSM-based approach. Masks can be changed to an arbitrary text.}
   \label{fig:targeted_attacks}
\end{figure}

Thanks to the changes we have made to the FGSM, we have influence over what mask will be generated during an attack. The aim of this algorithm is to minimize the distance between the generated mask and any target, so we also performed attempts to display text on the mask. Example results can be seen in Figure \ref{fig:targeted_attacks}. Unfortunately, generating this type of attack requires many iterations, so we performed these tests on a smaller scale.

\subsubsection{Black-box attacks}

We carried out analogous experiments for black-box methods. As can be seen in Table \ref{tab:main_results}, SIMBA outperformed EBAD in our tests despite the higher visibility of perturbations. However, the quantitative results clearly indicate that both methods fail to produce successful attacks, showcasing the difficulties black-box methods encounter with complex models like SAM.

Interestingly, EBAD is unable to achieve a mask different from the original (on average 96\% of IoU) regardless of the prompt used. We hypothesize that the implicit assumptions of the algorithm are the main cause since the ensemble should consist of models similar to the victim model. However, models from the original experiments, which we also use, possess significantly fewer parameters than SAM (e.g. 44.5M of ResNet101 \cite{zagoruyko2017wide} vs. 636M of SAM).

\subsubsection{Robustness of attacks}

\begin{figure}[h]
  \centering
  \includegraphics[width=0.9\linewidth]{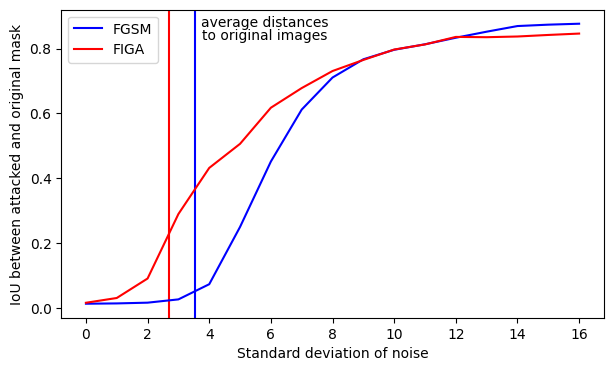}
  \caption{Performance of FGSM and FIGA attacks measured against added noise with increasing standard deviation. IoU between the attacked mask and the original mask increases when adding noise with greater standard deviation. The vertical lines show what is the average distance from the adversarial example to the original image (for both methods separately).}
  \label{fig:robustness}
\end{figure}

We have already shown that FGSM and FIGA can generate effective attacks by making only small changes that are invisible to the human eye. It is then worth seeing how robust the examples achieved at such a small cost are. To test this, we generated random Gaussian zero-centered noise with increasing standard deviation for all non-targeted attacks from our prepared dataset. One might have expected that perturbing the attacks would clearly weaken their effectiveness since the aim of both white-box methods is to find a local minimum beyond which a good model should behave normally. This conjecture coincides with our results for the FIGA algorithm, as disrupting the images caused a sudden drop in attack quality. However, the results we obtained for FGSM show that even after introducing a small noise, the attacks are effective, which can be seen in Figure \ref{fig:robustness}.

\begin{figure}[h]
  \centering
  \includegraphics[width=0.999\columnwidth, trim={2.6cm 23cm 2.7cm 0cm}, clip]{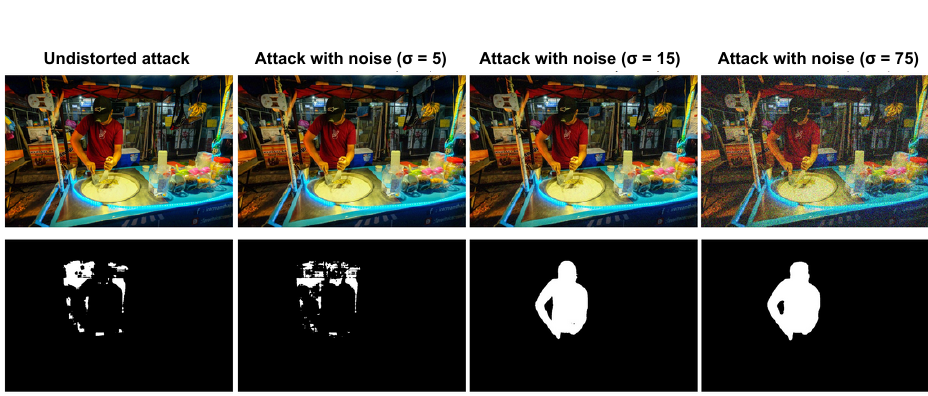}
  \caption{Behaviour of the FGSM attack when adding noise with increasing standard deviation. Big values of standard deviation prevent the attack from destroying the mask.}
  \label{fig:noise}
\end{figure}

The average distance (in $L_2$ norm) of the adversarial example generated using FGSM from the original image is only 3.6. When the Gaussian noise of this standard deviation is added, the average attack is still successful, which means that the random noise is not an effective defense. Only the introduction of large noise restores proper segmentation which also shows that SAM can robustly segment even noisy images. An example of this can be seen in Figure \ref{fig:noise}.

\section{Defense strategies}
We hypothesize that the performance of SAM on images of roads modified with style transfer could be improved by fine-tuning SAM with specially augmented photos \cite{corruptions_survey}. Undesired memorization of celebrities' faces could be due to features appearing in large datasets and be potentially resolved by better filtering of the datasets before training, identifying neurons that hold most of such information \cite{neuron_shapley} and resetting them or other unlearning techniques \cite{machine_unlearning}.
Finally, we believe that robustness against white-box attacks could be largely increased by adversarial training \cite{pgd} or fine-tuning the model in a special adversarial setup suited for segmentation tasks \cite{robust_segmentation}. Future research can focus on analyzing and applying such defense techniques to make SAM more robust.

\section{Conclusion}
\label{sec:conclusion}
Segment Anything is a useful foundation model used as a component in many
complex pipelines solving real-world problems. Throughout this paper, we argue that one
has to be careful when using SAM and rely on explainability analysis to assess what are the strengths and weaknesses of this model in specific tasks. For example, we demonstrated that SAM is robust to changing weather conditions unless they get very extreme and feature big distortions in images caused by the presence of water drops. 

Moreover, we show that the model's embeddings created by LangSAM hold specific details about certain celebrities. It becomes very concerning due to the potential for misuse and invasion of peoples' privacy. Our study highlights the effectiveness of white-box attack methods in identifying vulnerabilities in SAM. It underscores the challenges in attacking large-scale models without access to their internal parameters, pointing to the need for robust defenses of machine learning systems. We believe the analysis shown in this paper could be repeated on subsequent segmentation models to evaluate their robustness.

\section{Acknowledgments}

The work of Bartlomiej Sobieski on this project was financially supported by the SONATA BIS grant 2019/34/E/ST6/00052 funded by the Polish National Science Centre (NCN). Krzysztof Jankowski and Mateusz Kwiatkowski were financially supported by the IDUB program at the University of Warsaw.

{
    \small
    \bibliographystyle{ieeenat_fullname}
    \bibliography{main}
}

\end{document}